\documentclass[letterpaper]{article} 
\usepackage[]{aaai23}  
\usepackage{times}  
\usepackage{helvet}  
\usepackage{courier}  
\usepackage[hyphens]{url}  
\usepackage{graphicx} 
\usepackage{natbib}  
\usepackage{caption} 
\usepackage{algorithm}
\usepackage{algorithmic}

\usepackage{newfloat}
\usepackage{listings}
\usepackage{xspace}
\usepackage{xcolor}
\usepackage{multirow}
\usepackage{amsmath}

\usepackage{booktabs}
\usepackage{amsthm}

\DeclareCaptionStyle{ruled}{labelfont=normalfont,labelsep=colon,strut=off} 
\floatstyle{ruled}
\newfloat{listing}{tb}{lst}{}
\floatname{listing}{Listing}
\title{Selective Knowledge Distillation\\for Non-Autoregressive Neural Machine Translation}
\title{Selective Knowledge Distillation\\for Non-Autoregressive Neural Machine Translation}
\author {
    Min Liu$^{1 *}$,
    Yu Bao$^{2}$,
    Chengqi Zhao$^{2}$,
    Shujian Huang$^{1,3}$
}
\affiliations {
    $^{1}$National Key Laboratory for Novel Software Technology, Nanjing University\\ $^{2}$ByteDance AI Lab\\
    $^{3}$Collaborative Innovation Center of Novel Software Technology and Industrialization\\
    {minliu@smail.nju.edu.cn}, 
    {huangsj@nju.edu.cn},
    {\{baoyu.001,zhaochenqi.d\}@bytedance.com}
}

\begin{document}

\maketitle

\renewcommand{\thefootnote}{\fnsymbol{footnote}}
\footnotetext[1]{Work done during the internship at ByteDance AI Lab.}
\renewcommand{\thefootnote}{\arabic{footnote}}

\begin{abstract}
Benefiting from the sequence-level knowledge distillation, the Non-Autoregressive Transformer (NAT) achieves great success in neural machine translation tasks. 
However, existing knowledge distillation has side effects, such as propagating errors from the teacher to NAT students, which may limit further improvements of NAT models and are rarely discussed in existing research.
In this paper, we introduce selective knowledge distillation by introducing an NAT evaluator to select NAT-friendly targets that are of high quality and easy to learn.
In addition, we introduce a simple yet effective progressive distillation method to boost NAT performance. 
Experiment results on multiple WMT language directions and several representative models show that our approach can realize a flexible trade-off between the quality and complexity of training data for NAT models, achieving strong performances.
Further analysis shows that distilling only 5\% of the raw translations can help an NAT outperform its counterpart trained on raw data by about 2.4 BLEU.
\end{abstract}

\section{Introduction}
Non-autoregressive Transformer~\cite[NAT,][]{gu2018non} introduces a promising paradigm of parallel decoding. 
Unlike sequentially predicting the words in an autoregressive model, NAT models can generate a sentence in parallel based on a conditional independence assumption, improving the inference speed by over 10 times. 
Besides, such a parallel decoding paradigm also has the potential to avoid the \textit{exposure bias} that has a long-term discussion in sequential decoding models \cite{vaswani2017attention}.
As a result, we see NAT models achieve great success in machine translation tasks \cite{qian-etal-2021-volctrans}, surpassing many autoregressive models in WMT21\footnote{http://statmt.org/wmt21/}.

\setcounter{footnote}{0}
Despite the great potential of NAT models, they rely on sequence-level knowledge distillation~\cite[KD,][]{kim2016sequence} to achieve success.
The introduced conditional independence assumption prevents NAT models from leveraging the inherent structures to overcome the \textit{multi-modality problem}, where each input may correspond to several valid outputs in the training data.
In such background, \citet{gu2018non} introduce sequence-level knowledge distillation to bypass the multi-modality problem of NAT models. 
They first train an autoregressive Transformer~\cite[AT,][]{vaswani2017attention} as a teacher model, and then train the NAT models using the teacher's output as targets.
The deterministic outputs generated by the teacher can directly avoid the one-to-many situation in raw training data and improve the performance of an NAT model by over 5.0 BLEU~\cite{bleu2002} in machine translation.

\begin{figure}[t]
    \centering
    \includegraphics[width=0.80\columnwidth]{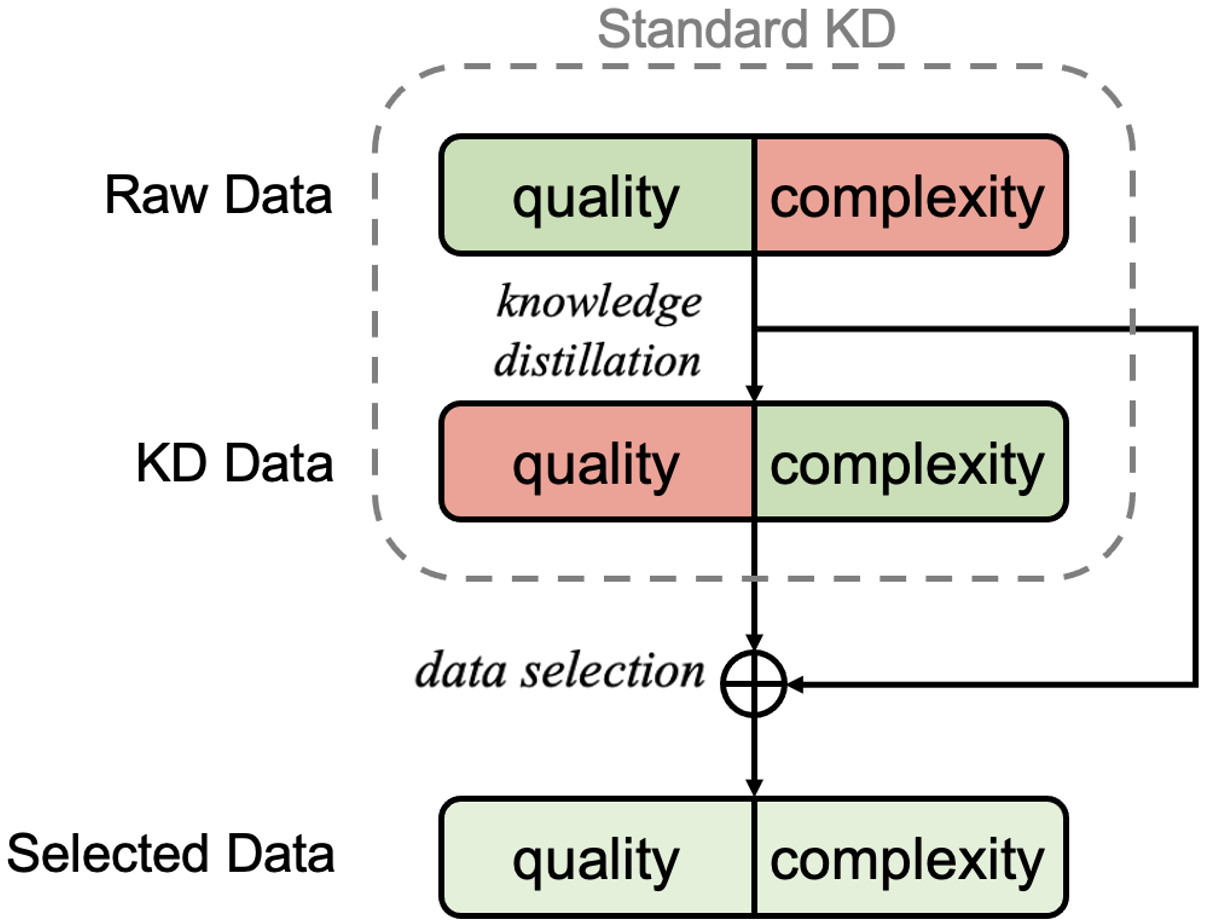}
    \caption{An illustration of our selective knowledge distillation. 
    Standard knowledge distillation reduces the complexity of raw data at the cost of translation quality. 
    In contrast, we propose combining the merits of raw and KD data, balancing the complexity and quality of training data.
    }
    \label{figure1}
\end{figure}


\begin{table*}[t]
\centering
\renewcommand{\arraystretch}{1.2}
\begin{tabular}{c|c|l}
    \toprule
     \textbf{\#} & \textbf{raw/distilled} & \textbf{outputs} \\
    \midrule
    \multirow{2}{*}{Translation \#1} & raw & I \textbf{entirely} agree with the PPE Group that paragraph 9 is \textbf{central} .\\
    & distilled & I \textit{fully} agree with the PPE Group that paragraph 9 is \textit{of key importance} .\\\hline
    \multirow{2}{*}{Translation \#2} & raw & That is up to the Heads of Governments to do \textbf{this} week . \\
    & distilled & That is up to the Heads of Government to do \textit{this this} week . \\\hline
    \multirow{2}{*}{Translation \#3} & raw & \textbf{Once you have started reading , you can not put it down .} \\
    & distilled & Anyone who starts reading it keeps his breath until the last word .\\\hline
    \multirow{2}{*}{Translation \#4} & raw & \textbf{Nice and clean hotel in great location , great value for money .} \\
    & distilled & Gravino Cinco is fresher and newer than Tryp Ciudad Hotel !\\
    \bottomrule
\end{tabular}
\caption{Examples of different situations. Our method mainly improves performance on sentences like Translation 2, where minor mistakes are introduced by the AT teacher. In our experiments, the NAT evaluator generated a translation exactly the same as the distilled one, while the NAT student trained on selected data corrected the mistake by removing the repeated token.}
\label{table1}
\end{table*}

However, there are still several problems in standard knowledge distillation, which may limit the performance of NAT models.
First, NAT models learning only from AT teachers may miss some important knowledge in the original data, such as prediction on low-frequency words~\cite{ding2021understanding}.
Second, the outputs generated by the AT teacher are not necessarily suitable for the training of NAT models, as these architectures have quite different modeling paradigms.
It should be noted that existing NAT research~\cite{gu2018non,ghazvininejad2019mask,guo2020jointly,bao2021non} only regards knowledge distillation as a necessary data processing technique but lacks a deeper discussion.
Therefore, designing knowledge distillation strategies to help NAT models learn better is still an open question.


In this paper, we propose a selective knowledge distillation technique for training NAT models to tackle the two issues in standard knowledge distillation.
More specifically, we introduce an NAT model trained on distilled data as an evaluator to construct the training data, replacing the original distilled data with raw data dynamically in the learning progress.
There are two intuitions behind our selective knowledge distillation:
First, our approach can access raw data and avoid repeating the mistakes made by the AT teacher.
Second, due to its similar modeling paradigm, the NAT evaluator can effectively assess whether the data is suitable for the training of NAT students.
The NAT evaluator judges each sentence in the original training set by scoring the predicted tokens.
We select sentences with higher scores as the targets which generally contain minor modality change from the distilled data but show better translation quality as raw data. In tuition, these sentences can be safely exposed to NAT students during training.
Besides, we introduce a hard-to-easy curriculum learning strategy while training, which has been demonstrated effective for automatic speech recognition systems \cite{braun2017curriculum}. 

We conduct experiments on two widely-used machine translation benchmarks, WMT14 En-De and WMT16 En-Ro and over an inference-efficient AT structure~\cite{kasai2020deep} and two representative NAT architectures~\cite{qian2021glancing, ghazvininejad2019mask}. 
Experiment results show that our selective knowledge distillation consistently improves models' performance on each dataset.
Further analyses show that a small ratio~(5\%) of distilled data is sufficient to improve NAT significantly, demonstrating that our method can effectively select the NAT-friendly raw translations. 
As an early attempt to introduce raw data for training NAT models, we hope this work will raise more attention to selecting beneficial examples from authentic data to recover the missing information while keeping the merits of knowledge distillation.

\section{Background}
Neural Machine Translation can be defined as a sequence-to-sequence generation problem: given source sentence $X=\{x_1, x_2,\cdots,x_N\}$, to generate target sentence $Y = \{y_1, y_2, \cdots, y_L\}$ according to $P(Y|X,\theta)$, where $\theta$ denotes the parameters of a network.

\subsection{Non-Autoregressive Neural Machine Translation}
Non-Autoregressive Transformer~\cite[NAT,][]{gu2018non} imposes the conditional independence assumption among target words while factorizing the probability $P(Y|X;\theta)$:
\[
    P(Y|X,\theta) = P(L|X, \theta)\prod_{i=1}^LP(y_i | X, \theta)
\]
where $L$ is the length of the target sequence. 


The conditional independence assumption allows NAT to significantly outperform autoregressive Transformer (AT) in inference speed, but it also leads to an inferior translation quality compared to AT.
A well-recognized explanation is that NAT models suffer from the multi-modality problem~\cite{gu2018non}, where the model fails to capture the highly multimodal distribution of target translations adequately. 
For example, a source sentence might have several ground-truth translations that differ in wording and structure, and NAT models are likely to get confused since they have to select from multiple choices only through the source sentence.
In contrast, an AT model can easily learn these different translations by predicting tokens based on the source sentence and previous tokens.

\subsection{Knowledge Distillation}
To alleviate the multi-modality problem, sequence-level knowledge distillation~\cite[KD,][]{kim2016sequence} is adopted as a preliminary step for training an NAT model, where the original translations are replaced with those generated by a pretrained autoregressive teacher. 
The distilled data eases the training by introducing more deterministic knowledge and significantly improves the performance of an NAT student.
Some previous works propose generating several distilled translations and select the most suitable candidate~\cite{zhou2019understanding,shao2022one} to gain more benefits from knowledge distillation.

However, knowledge distillation has some side effects like leading to more errors on low-frequency words \cite{ding2021understanding}. 
Due to the differences in architectures between AT and NAT, the translations generated by an AT teacher are not always suitable for the learning of NAT.
Another obvious limitation of KD is that it propagates mistakes made by the AT teacher to NAT students. 
Table \ref{table1} shows that the translation generated by the AT teacher contains mistakes which might harm the performance of the student. 
Therefore, how to break such limitations of AT-based KD and utilize authentic data to improve the translation quality remains a question.

\section{Method}
The intuition behind our method is that introducing raw translations which do not significantly increase the complexity will not make training much more challenging but free NAT from some mistakes made by the AT teacher. 
Section~$\S$\ref{ss:selecting} introduces how to select NAT-friendly raw sentences, combining the high translation quality of raw data and reduced complexity of distilled data. 
Besides, we also introduce a hard-to-easy learning strategy for dynamically configuring the raw data ratio in the training process, as presented in Section~$\S$\ref{ss:hard}.

\subsection{Selecting NAT-friendly Raw Translations}\label{ss:selecting}
While raw data are of high quality in most cases, the multi-modality problem prevents an NAT model from capturing the distribution of target translations properly. Contrarily, distilled data eases the training of NAT by reducing the complexity of targets, but mistakes made by the AT teacher will be easily propagated to the student if only distilled data is exposed. It is natural to think that NAT models should learn some missing information in the distilled translations from the original ones to improve translation quality, and a simple solution is to expose part of the raw data to NAT. The question remaining is how to evaluate whether a raw translation should be exposed.

We propose to evaluate each translation in the raw data through an NAT evaluator trained on distilled data, replacing a raw translation with its distilled version when the NAT evaluator fails to generate outputs similar to the reference. 
Specifically, given source sentence $X$, we first get a decoded output $\hat{Y}=f_{teacher}(X)$ using the NAT evaluator. 
Then we evaluate the raw translation $Y$ through a metric $score(X, Y)=1-d(Y, \hat{Y}) / |Y|$, which measures the difference between the ground truth translation and the predicted output. The translations with high scores are considered NAT-friendly.

In the following part we explain why an NAT evaluator can decide whether a raw sentence can be safely exposed. 
Since we want to keep both the high translation quality of raw sentences and the simplified modes of distilled sentences, a naive answer is that raw sentences with fewer modes can be set as targets for NAT. 
If the NAT evaluator trained on distilled data can get a prediction close to the raw target, then the raw and distilled translations are probably quite similar in their modes. 
To illustrate the details, here we list four typical situations where the distilled translations are different from the original ones:
\begin{itemize}
    \setlength\itemsep{0em}
    \item \textbf{Minor Modality Change:} A few words are substituted by their synonyms without introducing great changes to the structure and semantics of the raw sentence. Therefore, both the raw and distilled translation can be used as the target.
    \item \textbf{Minor Mistakes:} While the structure and semantics of the original sentence is preserved, a few mistakes like falsely predicted low-frequency words or word repetition are introduced. Learning from the raw translation can be helpful to correct these mistakes.
    \item \textbf{Dramatic Modality Change:} Despite sharing the same semantics, the raw and distilled translation are expressed in quite different ways. The raw translation contains modes too challenging for an NAT model.
    \item \textbf{Dramatic Mistakes:} The distilled sentence is not well-translated, but we are not sure whether the raw translation is a better target to learn from since even the AT.
\end{itemize}
Table~\ref{table1} provides the examples corresponding to each situation.
Minor modality changes~(Translation \#1) can be tolerated since they do not greatly increase the modes of training data, and correcting minor mistakes~(Translation \#2) is the main goal of our method.
The NAT evaluator is not likely to get a close prediction when there exists dramatic differences between raw and distilled data~(Translation \#3), so when it gives a raw sentence a high score, it is highly likely that the sentence satisfies our requirement of simple and clean translation. 
Besides, an NAT evaluator can avoid the cases where a distilled sentence is close to the original one but still too challenging for an NAT~(Translation \#4). 
Therefore, we can choose to distill only the raw sentences with low scores under the NAT evaluator and keep the rest unchanged. 
In this way, the dataset displays higher translation quality while keeping the general complexity suitable for NAT.

\begin{algorithm}[t]
    \caption{Data Selection for the $k$-th Update}
    \label{algorithm1}
    \begin{algorithmic}[1]
        \REQUIRE $\mathcal{D}_k\{(X, Y, Y^{KD})\}$, NAT evaluator $f_{teacher}$
        \STATE{$T_k\leftarrow T_0 + k / K\cdot (T_1 - T_0)$}
        \STATE $\mathcal{D}_k' \leftarrow \{\}$
        \FORALL{$(X, Y, Y^{KD})\in\mathcal{D}_k$}
            \STATE $score(X, Y)\leftarrow 1-d(Y, f_{teacher}(X))/|Y|$
            \IF{$score(X, Y) \geq T_k$}
                \STATE $\mathcal{D}_k'\leftarrow \mathcal{D}_k'\cup \{(X, Y)\}$
            \ELSE
                \STATE $\mathcal{D}_k'\leftarrow \mathcal{D}_k'\cup \{(X, Y^{KD})\}$           
            \ENDIF
        \ENDFOR
        \RETURN $\mathcal{D}_k'$
    \end{algorithmic}
\end{algorithm}

\subsection{Hard-to-Easy Data Selection}\label{ss:hard}
Motivated by the success of curriculum learning~\cite{qian2021glancing,fcl_nat,tcl_nat}, we further introduce a hard-to-easy learning strategy to improve the performance. 
\citet{ding2021understanding} show that pretraining with raw data can improve the performance of NAT by rejuvenating low-frequency words. 
To keep the merits of low-mode, they further trained the pretrained model on distilled data. We combine this idea with our data selection method by decreasing the ratio of raw data in the training process. Specifically, the training data for each update can be formulated as:
\[
\begin{split}
\{ (X, Y) | score(X, Y) \geq T_k \land (X, Y, Y^{KD}) \in \mathcal{D}_k \} \cap\\
\{ (X, Y^{KD}) | score(X, Y) < T_k \land (X, Y, Y^{KD}) \in \mathcal{D}_k \}
\end{split}
\]
where $T_k$ and $\mathcal{D}_k$ denote the threshold and the set of tuples $(X,Y,\hat{Y})$ for the $k$th update respectively.
$T_k$ can be determined by a preset function or feedbacks from the NAT student. 
In our experiments, we adopt a linear function for $T_k$ which is computed as $T_k=T_0 + \frac{k}{K}(T_1 - T_0)$, where $K$ is the total number of updates, the constants $T_0$ and $T_1$ can be determined according to the distribution of score $P(score(X,Y))$ given a specific NAT evaluator and the raw training data. 
The whole data selection process can be found in Algorithm \ref{algorithm1}. 
This process is an additional stage following standard training procedures for NAT, thus being generic to various data and architectures.

\begin{table*}[t]
    \centering
    \renewcommand{\arraystretch}{1.05}
    \begin{tabular}{lccccc}
    \toprule[1.2pt]
        \multicolumn{1}{c}{\multirow{2}{*}{\textbf{Methods}}} & \multirow{2}{*}{\textbf{Iter}} & \multicolumn{2}{c}{\textbf{WMT14}} & \multicolumn{1}{c}{\textbf{WMT16}}  & \multirow{2}{*}{\textbf{Speed Up}} \\\cline{3-5}
        & & \textbf{En-De} & \textbf{De-En} & \textbf{En-Ro}\\\hline
        \multicolumn{6}{c}{\textbf{AT Models}}\\
        \textbf{Transformer} \cite{vaswani2017attention} & T & 27.30 & / & / & 1.0$\times$ \\
        \textbf{Transformer} * & T & 27.34 (0.309) & 31.73 (0.388) & 34.68 (0.515) & 1.0$\times$ \\
        \textbf{DeepShallow} \cite{kasai2020deep} * & T & 26.00 (0.152) & 30.62 (0.308) & 32.25 (0.401) & 2.4$\times$ \\\hline
        \multicolumn{6}{c}{\textbf{Iterative NAT Models}}\\
        \textbf{CMLM} \cite{ghazvininejad2019mask} & 10 & 27.03 & 30.53 & 33.08 & 1.7$\times$\\
        \textbf{JM-NAT} \cite{guo2020jointly} & 10 & 27.31 & 31.02 & / & 5.7$\times$\\
        \hline
        \multicolumn{6}{c}{\textbf{Non-iterative NAT Models}}\\
        \textbf{NAT-FT} \cite{gu2018non} & 1 & 17.69 & 21.47 & 27.29 & 15.6$\times$ \\
        \textbf{GLAT} \cite{qian2021glancing} & 1 & 25.21 & 29.84 & 31.19 & 15.3$\times$ \\
        \textbf{GLAT + CTC} \cite{qian2021glancing} & 1 & 26.39 & 29.54 & 32.79 & 14.6$\times$ \\
        \textbf{DA-Transformer} \cite{pmlr-v162-huang22m} & 1 & 27.91 & 31.95 & / & 7.0$\times$\\
        \hline
        
        \multicolumn{6}{c}{\textbf{Our Models}}\\
        \textbf{DeepShallow w/ Standard KD} * & T & 27.05 (0.246) & 31.36 (0.326) & 32.99 (0.416) & 2.4$\times$ \\
        \textbf{DeepShallow w/ Selective KD~(ours)} & T & \textbf{27.23} (\textbf{0.252}) & \textbf{31.70} (\textbf{0.352}) & \textbf{33.28} (\textbf{0.438}) & 2.4$\times$ \\
        \textbf{CMLM w/ Standard KD} * & 10 & 26.64 (0.137) & 30.24 (0.215) & 32.85 (0.357) & 2.1$\times$ \\
        \textbf{CMLM w/ Selective KD~(ours)} & 10 & \textbf{27.06} (\textbf{0.170}) & \textbf{30.65} (\textbf{0.226}) & \textbf{33.38} (\textbf{0.374}) & 2.1$\times$ \\
        \textbf{GLAT + CTC w/ Standard KD} * & 1 & 26.19 (0.119) & 30.74 (0.274) & 32.73 (0.362) & 14.2$\times$ \\
        \textbf{GLAT + CTC w/ Selective KD~(ours)} & 1 & \textbf{26.82} (\textbf{0.144}) & \textbf{31.30} (\textbf{0.302}) & \textbf{33.34} (\textbf{0.381}) & 14.2$\times$ \\
    \bottomrule[1.2pt]
    \end{tabular}
    \caption{BLEU and COMET scores of NAT models on WMT14 En-De/De-En and WMT16 En-Ro benchmarks. COMET scores are listed in parentheses if available. * indicates the results are obtained based on our implementation. To highlight the advantage in efficiency, we did not apply strategies like reranking which improve the performance at the cost of inference speed. }
    \label{table2}
\end{table*}

\section{Experiments}
\subsection{Experimental Settings}
\paragraph{Datasets} We conduct experiments on two widely-used machine translation datasets: WMT14 English-German (En-De) and WMT16 English-Romanian (En-Ro), which consist of 3.96M and 0.6M sentence pairs, respectively. 
Following the common practices, we process the datasets with Moses script \cite{koehn-etal-2007-moses} and segment the words into subword units using byte-pair encoding \cite[BPE,][]{sennrich2016neural}. 
The subword embeddings are shared between the source and target language. 
For the sequence-level knowledge distillation, we employ the Transformer with base settings in \citet{vaswani2017attention} as the teacher. 

\paragraph{Model} We evaluate our selective knowledge distillation on DeepShallow \cite{kasai2020deep}, CMLM \cite{ghazvininejad2019mask}, and GLAT+CTC \cite{qian2021glancing}.
DeepShallow is an inference-efficient AT structure with a deep encoder and a single-layer autoregressive decoder, which also benefits from knowledge distillation. We adopt a 6-layer encoder in the experiments.
CMLM iteratively generates the target sequence from the masked input.
For the previous two models, we compute $d(Y, \hat{Y})$ using the Hamming distance $\sum_{i=1}^{L}[Y_i\neq \hat{Y}_i]$. 
GLAT builds the word interdependencies to improve the performance of single-pass parallel generation. 
During training, the decoder is fed with randomly masked target sequence, and the number of masked tokens depends on the prediction accuracy. 
The performance of GLAT can be further improved by connectionist temporal classification \cite[CTC,][]{graves2006connectionist}, which utilizes an alignment-based objective. Another advantage of CTC is that it can align the targets according to decoder outputs so that ground-truth tokens are not required to be predicted on a fixed position, thus making the NAT evaluator more tolerant to minor mistakes when evaluating a raw translation. 
To compute $score(X, Y)$ for applying our approach on GLAT+CTC, we use dynamic programming to get the aligned path $Y^{align}$ with the largest align score \cite{graves2006connectionist} and adopt the Hamming distance as the metric, which is computed as $d(Y^{align}, \hat{Y})=\sum_{i=1}^{L'}[Y^{align}_i\neq \hat{Y}_i]$. 

\paragraph{Training Settings} 
We follow the hyperparameters of models in their original papers. 
We set the dropout rate to $0.1$ for WMT14 En-De/De-En and $0.3$ for WMT16 En-Ro. 
For the optimizer, we use Adam with $\beta=(0.9, 0.999)$ to train our model. 
The learning rate warms up to $5e-4$ within 4k steps and then decays with the inverse square-root schedule. 
For the sampling ratio $\lambda$ in GLAT+CTC, we adopt linear annealing from 0.5 to 0.3. 
As to the hard-to-easy learning strategy, we set $T_0=0.4, T_1=1.0$ under En-De/De-En and $T_0=0.6, T_1=1.0$ under En-Ro for GLAT+CTC. We set $T_0=0, T_1=1.0$ for other models.
All the NAT evaluators and students are trained with batches of 64k tokens, lasting 300k updates and 100k updates for En-De/De-En and En-Ro respectively. 
To better utilize the NAT evaluators, the students are initialized with parameters of the teachers trained after 25k updates for En-De/De-En and 10k updates for En-Ro, when the general knowledge has been acquired. 
We average the top 5 checkpoints chosen by the validation BLEU scores to create the final model. 

\begin{figure*}[t]
    \centering
    \includegraphics[width=0.85\columnwidth]{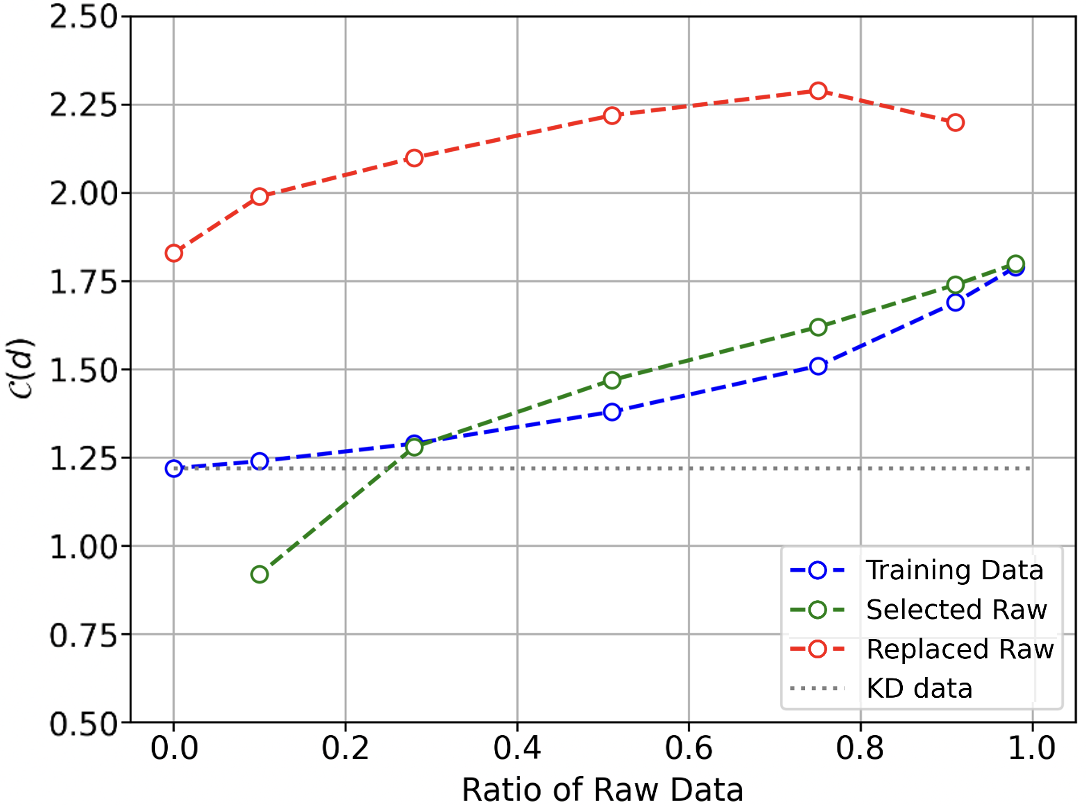}
    \includegraphics[width=0.85\columnwidth]{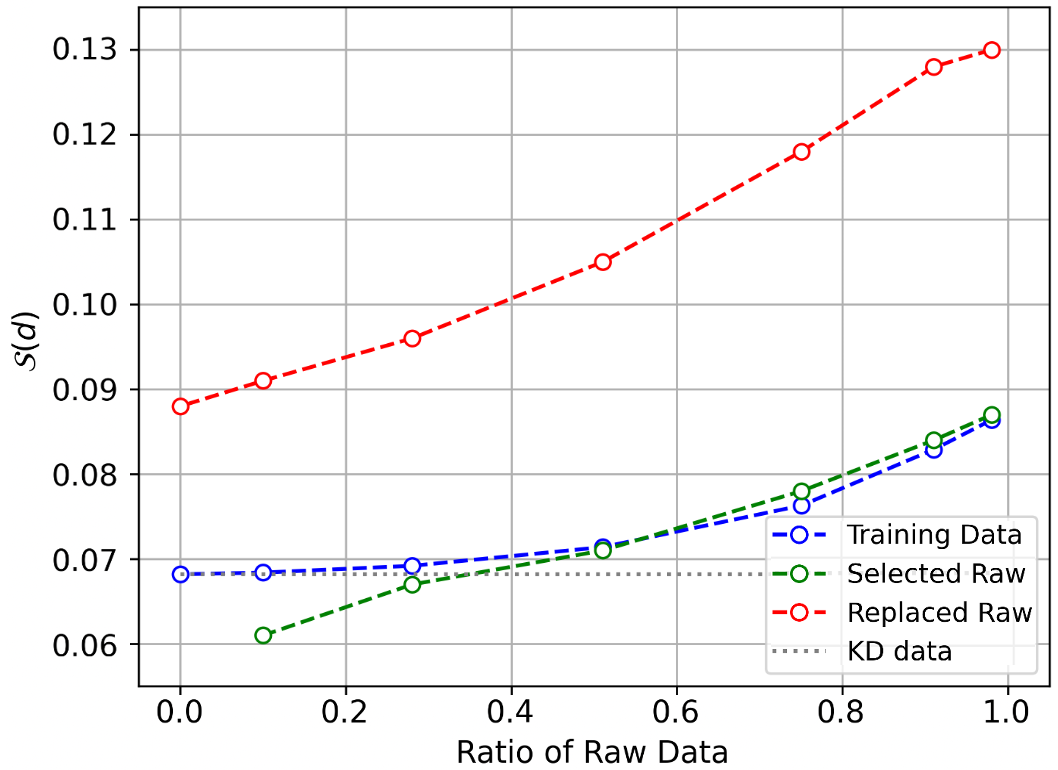}
    \caption{$C(d)$ and $S(d)$ on 1M pairs randomly sampled from WMT14 En-De. We set $T=[0.4, 0.5, 0.6, 0.7, 0.8, 0.9, 1.01]$ for the experiments, and the ratio of raw data=$[0.98, 0.91, 0.75, 0.51, 0.28, 0.10, 0.00]$ respectively with a GLAT+CTC evaluator. \textit{Selected Raw} is the set of raw sentences selected under $T$, while \textit{Replaced Raw} is the set of raw sentences to be distilled. We concatenate \textit{Replaced Raw} after distillation and \textit{Selected Raw} to get \textit{Training Data}, which is the data exposed to the NAT student during training. We neglect $C(d)$ and $S(d)$ when there is not enough data for analysis.}
    \label{figure2}
\end{figure*}

\paragraph{Baselines} We compare our method with standard KD which distills the whole training set. 
Another baseline is Low Frequency Rejuvenation~\cite[LFR,][]{ding2021rejuvenating}, which also exposes raw data to the NAT. 
They trained NAT models with raw, bidirectional KD and standard KD data in three different stages.
We also apply their method to GLAT+CTC with the training updates split to approximately $2:2:3$ in ratio for each stage. 
Their method is trained for 325k updates on En-De/De-En and 110k updates on En-Ro for fair comparison. 
Note that their method augments the training data by introducing \textit{(distilled source, raw target)} sentence pairs, while ours only utilizes raw and standard KD data. We evaluate all the models using the tokenized and cased BLEU scores \cite{bleu2002}, and a learned metric COMET \cite{rei2020comet} with the recommended model wmt20-comet-da.

\begin{table}[t]
    \centering
    \renewcommand{\arraystretch}{1.0}
    \begin{tabular}{cccc}
        \toprule[1.2pt]
        \multirow{2}{*}{\textbf{Methods}} & \multicolumn{2}{c}{\textbf{WMT14}} & \textbf{WMT16} \\\cmidrule{2-4}
        & \textbf{En-De} & \textbf{De-En} & \textbf{En-Ro} \\\midrule
        \textbf{LFR} & 26.56 & 31.13 & 33.27 \\
        \textbf{Selective KD~(ours)} & 26.82 & 31.30 & 33.34 \\
        \bottomrule[1.2pt]
    \end{tabular}
    \caption{BLEU scores of GLAT+CTC using our method and LFR~\cite{ding2021rejuvenating} based on our implementation.}
    \label{table_lrf_ours}
\end{table}

\subsection{Main Results}
Table \ref{table2} and Table \ref{table_lrf_ours} present the main results on the benchmarks.
Our method outperforms baselines consistently across different language pairs. 
We enable the model to learn directly from authentic data without greatly increasing the modes by selecting NAT-friendly raw translations using an NAT evaluator. 
Compared with the previous work \cite{ding2021rejuvenating} which also exposes raw data directly to NAT, we can determine the period of exposure for each sentence by setting the threshold dynamically in the training process. 
We highlight the empirical advantages of our method:
\begin{itemize}
\setlength\itemsep{0em}
    \item Simple, effective and generic. Our method adds a simple data selection procedure to the standard training pipeline, while it can effectively improve the performance of NAT across different datasets. Since the method is architecture-irrelevant, it can be applied to a wide range of architectures while maintaining their advantages, even including inference-efficient AT structures. 
    \item Well balance the translation quality and complexity of data.
    Our method can configure the translation quality and complexity of training data by setting different thresholds for data selection. As the ratio of raw data increases, the translation quality improves, and the complexity of training data increases only slightly since we deliberately select the simple raw translations.
\end{itemize}

\subsection{Analysis}
\paragraph{Properties of Selected Raw Data.}
Our method aims at selecting more NAT-friendly raw translations, which contain few modes and show high quality. To validate that our data selection process indeed find a set of training data that has the desired properties, we measure the complexity of our training data using two metrics:
\vspace{-0.2em}
\begin{itemize}
    \item \textbf{Translation Uncertainty}: \citet{zhou2019understanding} proposed to measure the translation uncertainty of parallel data based on conditional entropy. They simplified conditional entropy to the sum of entropy of target words conditioned on the aligned source words:
    \[
        C(d) = \frac{1}{|\mathcal{V}_x|}\sum_{x\in\mathcal{V}_x} \mathcal{H}(y|x)
    \]
    where $d$ is a given dataset and $\mathcal{V}_x$ is the set of source vocabularies. 
    \vspace{-0.2em}
    \item \textbf{Alignment Shift}: We measure the change of sentence structure according to the relative distance between aligned words. Specifically, given source sentence $X$ and its translation $Y$, we get 
    \[
        \tau(X, Y)=\frac{1}{|Y|}\sum_{i,j}[X_i = \text{align(}Y_j\text{)}]\cdot|\frac{i}{|X|}-\frac{j}{|Y|}|.
    \]
    $S(d)$ is computed as the average of $\tau(X, Y)$ over all pairs: $S(d) = \frac{1}{|d|}\sum_{(X, Y)\in d}\tau(X, Y)$.
    
\end{itemize}
\vspace{-0.2em}

We adopt an alignment model \cite{dyer2013simple} for the metrics above. The metrics are computed over 1M randomly sampled sentence pairs from our processed WMT14 En-De. To display the effects of our method, we compute the metrics for distilled data, selected raw data (using GLAT+CTC), raw data replaced by KD data and the overall training data under different threshold $T$.

As shown in Figure \ref{figure2}, the translation uncertainty and alignment shifts of replaced raw data (red) exceed those of selected raw data (green) by a large margin, indicating that our method can effectively separate raw data into classes of different complexity. When the threshold $T$ is high enough, the selected raw data even displays lower complexity than the average level of distilled data. 
This further proves that the selected raw data contains fewer modes. Observing the results on training data (blue), we find that the metrics grow smoothly as the ratio of raw data increases, which means that a flexible trade-off between translation quality and complexity of data can be realized.

\begin{figure}[t]
    \centering
    \includegraphics[width=0.76\columnwidth]{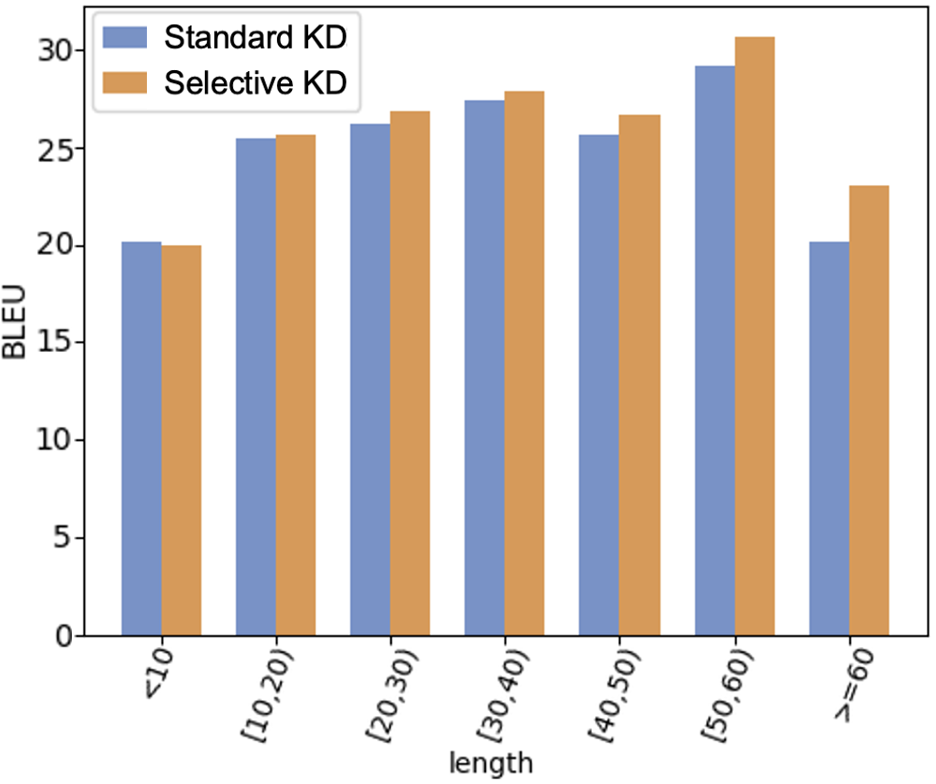}
    \caption{BLEU scores of GLAT+CTC for examples of different lengths on WMT14 En-De.}
    \label{figure3}
\end{figure}

\begin{table}[t]
    \centering
    \renewcommand{\arraystretch}{1.0}
    \begin{tabular}{cccc}
        \toprule[1.2pt]
        \textbf{Model} & \textbf{En-De} & \textbf{De-En} & \textbf{En-Ro} \\\hline
        \textbf{Standard KD} & 1.06\textperthousand & 0.56\textperthousand & 0.80\textperthousand \\
        \textbf{Selective KD} & 0.82\textperthousand & 0.38\textperthousand & 0.64\textperthousand \\
        \bottomrule[1.2pt]
    \end{tabular}
    \caption{Word repetition ratio of GLAT+CTC on WMT14 En-De/De-En and WMT16 En-Ro.}
    \label{table3}
\end{table}

\paragraph{Our Method Reduces Repetition.}
We also measure the percentage of repeated tokens to analyze whether our method can reduce the occurrence of repetition which is a typical mistake caused by the multi-modality problem. 
We see in Table \ref{table3} that exposing raw data during training can further reduce token repetition ratio.
Although our data contains more modes than fully distilled data, it still achieves a better result.
We think the improvement comes from learning directly from authentic distribution, which exhibits better word interdependencies and fewer mistakes.

\paragraph{Long Sentences Benefits More.} 
Figure \ref{figure3} presents the BLEU score on sentences of different lengths. 
As seen, longer sentences benefit more from our selective knowledge distillation.
Intuitively, the long sentences may contain more mistakes during distillation; thus, learning from authentic data can help the NAT student avoid or correct these mistakes and strengthen its ability to model long sentences. 
We also find that the performance drops slightly on sentences with fewer than ten tokens.
As shown in Table \ref{table4}, shorter sentences have higher average scores, thus exposed to the student NAT for a longer period. 
In such a case, long-term exposure to raw data may confuse the model's training, as it suffers from the multi-modality of the raw data.

\begin{table}[t]
    \centering
    \renewcommand{\arraystretch}{1.0}
    \begin{tabular}{ccc}
        \toprule[1.2pt]
        \textbf{Length} & \textbf{Score} & \textbf{Exposure Period} \\\hline
        $<10$ & 0.826 & 71.0\% \\
        $[10, 20)$ & 0.740 & 56.6\% \\
        $[20, 30)$ & 0.696 & 49.3\% \\
        $[30, 40)$ & 0.680 & 46.6\% \\
        $[40, 50)$ & 0.670 & 45.1\% \\
        $[50, 60)$ & 0.658 & 43.0\% \\
        $\geq 60$ & 0.644 & 40.6\% \\
        \bottomrule[1.2pt]
    \end{tabular}
    \caption{Average score and exposure period for raw translations of different lengths on WMT14 En-De with a GLAT +CTC evaluator. Exposure period is given by the percentage of updates where the raw translation can be directly learned.}
    \label{table4}
\end{table}

\begin{figure}[t]
    \centering
    \includegraphics[width=0.78\columnwidth]{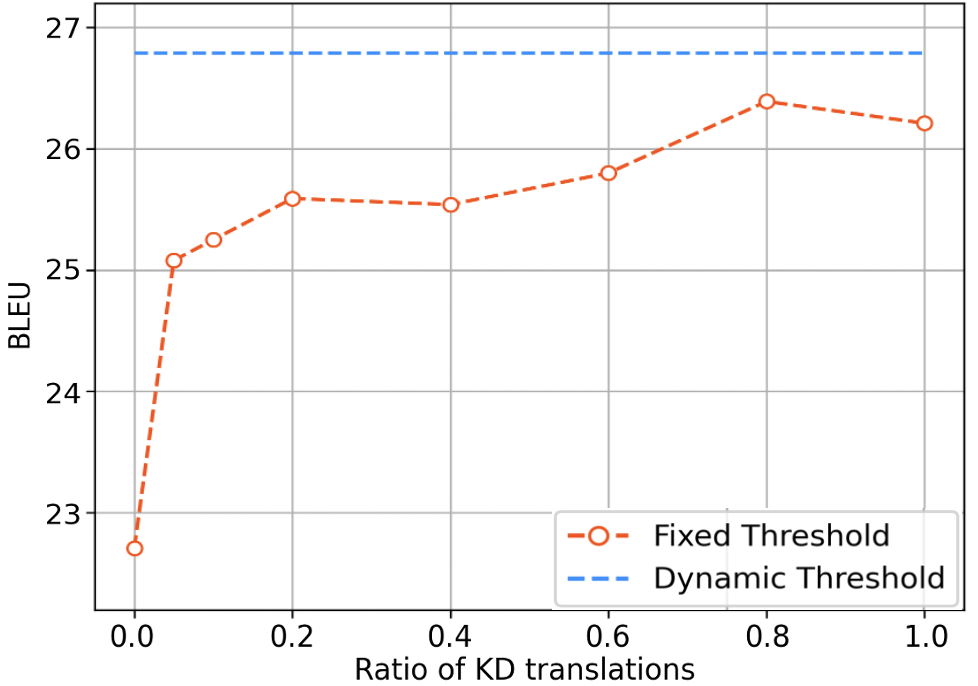}
    \caption{Performance of GLAT+CTC on WMT14 En-De with fixed threshold and dynamical threshold (0.4$\rightarrow$1.0).}
    \label{figure4}
\end{figure}

\subsection{Ablation Study}
\paragraph{Effects of Threshold $T$.}
We further analyze the effects of threshold $T$ in Figure \ref{figure4}.
We fix the threshold $T$ so that the training data remains unchanged during the training process. 
The model can achieve significant improvement (+2.4 BLEU) by distilling only 5\% of the training data. 
We attribute this phenomenon to the effectiveness of our data selection process, which can filter translations that greatly complicate the training data. 
The growth in performance becomes much slower as the ratio of distilled translations increases. 
Another finding is that the model trained on 80\%-distilled data slightly outperforms the one trained on fully distilled data. 
According to \citet{zhou2019understanding}, a potential explanation is that the complexity of the 80\%-distilled data is more suitable for the capacity of GLAT+CTC architecture. 
The dynamic threshold outperforms all the fixed threshold settings, embodying the advantage of our hard-to-easy strategy.

\paragraph{Model Initialization.} 
To study how model initialization influences our method, we initialize the GLAT+CTC student with parameters of the teacher trained after $t$ updates, where $t$ ranges from 25k to 300k with step 25k. 
We find that initialization with teacher trained after only 25k updates when the improvement on validation set begins to slow down achieves the best performance (26.82 BLEU), but the performance gap between these differently initialized models is negligible. This suggests that the improvement of our method does not come from a longer training process (initialization + training).
However, removing teacher initialization brings about a degeneration of 0.47 BLEU. 
We believe that transferring some basic knowledge from the teacher can free the student from learning everything from scratch on the more challenging raw data, enabling the student to focus on the missing knowledge in distilled data.

\section{Related Work} 
\paragraph{Non-autoregressive Machine Translation} \citet{gu2018non} first proposed Non-Autoregressive Transformer (NAT) for machine translation, which significantly boost the inference speed by generating the outputs in parallel. Despite the efficiency, NAT still lags behind AT in performance. Various methods have been proposed to bridge the performance gap. A line of work proposes to enhance the decoder inputs of NAT \cite{lee2018deterministic, wei2019imitation, wang2019non}. Another branch of work proposes to model the interdependencies between target outputs, which is explicitly missing in vanilla NAT \cite{ghazvininejad2019mask, qian2021glancing}. In addition, a series of work takes the latent variable as inputs to modeling the target-side information \cite{kaiser2018fast, ma2019flowseq, akoury2019syntactically, bao2021non, bao2022glat}. These work lines focus on model architecture and training method, so they can be easily combined with our model-agnostic method.

\paragraph{Training Data Manipulation} More close to our work is the thread of studies on manipulating training data for NAT. \citet{zhou2019understanding} show that sequence-level knowledge distillation \cite{kim2016sequence} reduces the complexity of training data and propose several methods to adjust the complexity of distilled data in order to match the model’s capacity. 
\citet{sun2020approach} jointly optimizes AT and NAT models to remove the multi-modality in target sentences. 
\citet{shao2022one} generate several high-quality reference translations and select the most suitable candidates by comparing them with the NAT outputs.
Some recent studies show that distilled data has some side effects like leading to more errors on predicting low-frequency words \cite{ding2021understanding}. In order to solve this problem, \citet{ding2021rejuvenating} proposed to pretrain NAT models on raw data, which is closely related to our work. Our method follows the idea of exposing raw data to NAT, but is different from theirs by introducing an NAT evaluator to evaluate each raw translation. By changing the ratio of raw sentences in the training data, we can configure the complexity of data in the training process and benefit more from raw data by exposing some raw translations for a longer period. 

\paragraph{Curriculum Learning} Our work adopts a hard-to-easy strategy in training NAT models by decreasing the ratio of raw data in the training process, which is contrary to curriculum learning \cite{bengio2009curriculum} in spirits. Curriculum learning methods train machine learning models from easy to hard data, but \cite{braun2017curriculum} showed that learning from hard to easy can be effective. They conducted experiments on automatic speech recognition systems and use signal-to-noise ratio (SNR) to create hard-to-easy curriculum. Compared with the opposite ranking of the examples from easy to hard, the hard-to-easy strategy provides better results.
 
\section{Conclusion}

In this paper, we propose selective knowledge distillation to tackle error propagation from an autoregressive teacher in standard knowledge distillation for NAT models.
Specifically, we employ an NAT evaluator to progressively replace the targets from distilled data with raw data for training NAT students, enabling them to benefit from both the high-quality raw data and easy-to-learn distilled data. 
Experiment results validate that our approach can effectively improve performance on machine translation tasks.
Extensive analyses also reveal that an effective data selection strategy has a great potential to improve the performance.


\section*{Acknowledgements}
We would like to thank the anonymous reviewers for their insightful comments. Shujian Huang is the corresponding author. This work is supported by National Science Foundation of China (No. 62176120), the Liaoning Provincial Research Foundation for Basic Research (No. 2022-KF-26-02).

\bibliography{aaai23}

\end{document}